\documentclass[11pt,a4paper]{article}
\usepackage{authblk}
\usepackage[hyperref]{acl2019}
\usepackage{times}
\usepackage{latexsym}

\usepackage{url}
\usepackage{graphicx}  %Required
\usepackage{tabularx}
\usepackage{multirow}
\usepackage{booktabs}
\usepackage{microtype}
\usepackage{tabu}
\usepackage[normalem]{ulem}
\usepackage[utf8]{inputenc}
\usepackage{algorithm}
\usepackage{amssymb}% http://ctan.org/pkg/amssymb
\usepackage{pifont}% http://ctan.org/pkg/pifont
\newcommand{\xmark}{\ding{55}}% 
\usepackage{algpseudocode}
\usepackage{array}
\usepackage{amsmath}
\usepackage{tablefootnote}
\usepackage{tikz}
\usetikzlibrary{calc,trees,positioning,arrows,chains,shapes.geometric,%
    decorations.pathreplacing,decorations.pathmorphing,shapes,%
    matrix,shapes.symbols,shadows,shapes.geometric}

\tikzset{
>=stealth',
  punktchain/.style={
    rectangle, 
    rounded corners, 
    draw=black, very thick,
    text width=10em, 
    minimum height=3em, 
    text centered, 
    on chain},
  line/.style={draw, thick, <-},
  element/.style={
    tape,
    top color=white,
    bottom color=blue!50!black!60!,
    minimum width=8em,
    draw=blue!40!black!90, very thick,
    text width=10em, 
    minimum height=3.5em, 
    text centered, 
    on chain},
  every join/.style={->, thick,shorten >=1pt},
  decoration={brace},
  tuborg/.style={decorate},
  tubnode/.style={midway, right=2pt},
startend1/.style={
    rectangle, 
    rounded corners=15pt, 
    text width=3cm, 
    minimum height=1cm,
    align=center, 
    line width=2pt,
    draw=white,
    font=\color{black}\sffamily, 
    fill=white
  }}

\usepackage{amsmath}
\usepackage{gb4e}

\makeatletter
\newcommand{\printfnsymbol}[1]{%
  \textsuperscript{\@fnsymbol{#1}}%
}
\newcommand{\corpusname}[0]{\textsc{KnowRef}}

\makeatother

\noautomath % need this; otherwise gb4e underscore issue crashes compilation

\usepackage{lipsum}

\newcommand\blfootnote[1]{%
  \begingroup
  \renewcommand\thefootnote{}\footnote{#1}%
  \addtocounter{footnote}{-1}%
  \endgroup
}

\newcolumntype{L}[1]{>{\raggedright\let\newline\\\arraybackslash\hspace{0pt}}m{#1}}
\newcolumntype{C}[1]{>{\centering\let\newline\\\arraybackslash\hspace{0pt}}m{#1}}
\newcolumntype{R}[1]{>{\raggedleft\let\newline\\\arraybackslash\hspace{0pt}}m{#1}}

\aclfinalcopy % Uncomment this line for the final submission
\title{Instructions for ACL 2019 Proceedings}

\author[1]{Ali Emami\textsuperscript{*}}
\author[1]{Paul Trichelair\textsuperscript{*}}
\author[2]{Adam Trischler}
\author[2]{Kaheer Suleman}
\author[2]{\\Hannes Schulz}
\author[1]{Jackie Chi Kit Cheung}
\affil[1]{School of Computer Science, Mila/McGill University}
\affil[2]{Microsoft Research Montreal}
\affil[ ]{\textit {\{ali.emami, paul.trichelair\}@mail.mcgill.ca}}
\affil[ ]{\textit {\{adam.trischler, kasulema, hannes.schulz\}@microsoft.com}}
\affil[ ]{\textit {jcheung@cs.mcgill.ca}}

\date{}

\begin{document}
\title{The \corpusname{} Coreference Corpus: Removing Gender and Number Cues for Difficult Pronominal Anaphora Resolution}

\maketitle
\begin{abstract}
\blfootnote{\textsuperscript{*}equal contribution}We introduce a new benchmark for coreference resolution and NLI, \corpusname{}, that targets common-sense understanding and world knowledge. Previous coreference resolution tasks can largely be solved by exploiting the number and gender of the antecedents, or have been handcrafted and do not reflect the diversity of naturally occurring text. We present a corpus of over 8,000 annotated text passages with ambiguous pronominal anaphora. These instances are both challenging and realistic. We show that various coreference systems, whether rule-based, feature-rich, or neural, perform significantly worse on the task than humans, who display high inter-annotator agreement. To explain this performance gap, we show empirically that state-of-the art models often fail to capture context, instead relying on the gender or number of candidate antecedents to make a decision. We then use problem-specific insights to propose a data-augmentation trick called \textit{antecedent switching} to alleviate this tendency in models. Finally, we show that antecedent switching yields promising results on other tasks as well: we use it to achieve state-of-the-art results on the GAP coreference task.
\end{abstract}

\section{Introduction}

%\blfootnote{*Equal contribution.}
Coreference resolution is one of the best known tasks in Natural Language Processing (NLP). Despite a large body of work in the area over the last few decades \cite{morton2000coreference,bean2004unsupervised,mccallum2005conditional,rahman2009supervised}, the task remains challenging. Many resolution decisions require extensive world knowledge and understanding common points of reference \cite{pradhan2011conll}. In the case of pronominal anaphora resolution, these forms of ``common sense'' become much more important when cues like gender and number do not by themselves indicate the correct resolution \cite{trichelair2018evaluation}.

To date, most existing methods for coreference resolution \cite{raghunathan2010multi,lee2011stanford,durrett2013decentralized,lee2017end,lee2018higher} have been evaluated on a few popular datasets, including the CoNLL 2011 and 2012 shared coreference resolution tasks \cite{pradhan2011conll,pradhan2012conll}. These datasets were proposed as the first comprehensively tagged and large-scale corpora for coreference resolution, to spur progress in state-of-the-art techniques.
According to \citet{durrett2013easy}, this progress would contribute in the ``uphill battle" of modelling not just syntax and discourse, but also semantic compatibility based on world knowledge and context. 

Despite improvements in benchmark dataset performance, the question of what exactly current systems learn or exploit remains open, particularly with recent neural coreference resolution models. \citet{lee2017end} note that their model does ``little in the uphill battle of making coreference decisions that require world knowledge,'' and highlight a few examples in the CoNLL 2012 task that rely on more complex understanding or inference. Because these cases are infrequent in the data, systems can perform very well on the CoNLL tasks according to standard metrics by exploiting surface cues. High-performing models have also been observed to rely on social stereotypes present in the data, which could unfairly impact their decisions for some demographics \cite{zhao2018gender}.

There is a recent trend, therefore, to develop more challenging and diverse coreference tasks. Perhaps the most popular of these is the Winograd Schema Challenge (WSC), which has emerged as an alternative to the Turing test~\cite{levesque2011winograd}.
The WSC task is carefully controlled such that heuristics involving syntactic salience, the number and gender of the antecedents, or other obvious syntactic/semantic cues are ineffective. Previous approaches to common sense reasoning, based on logical formalisms \cite{bailey2015winograd} or deep neural models \cite{liu2016probabilistic}, have solved only restricted subsets of the WSC with high precision. These shortcomings can in part be attributed to the limited size of the corpus (273 instances), which is a side effect of its hand-crafted nature.  \citet{webster2018mind} recently presented a corpus called GAP that consists of about 4,000 unique binary coreference instances from English Wikipedia. This corpus is intended to address gender bias and the mentioned size limitations of the WSC. We believe that gender bias in coreference resolution is part and parcel of a more general problem: current models are unable to abstract away from the entities in the sentence to take advantage of the wider context to make a coreference decision.

To tackle this issue, we present a coreference resolution corpus called \corpusname{} that specifically targets the ability of systems to reason about a situation described in the context.\footnote{The corpus, the code to scrape the sentences from the source texts, as well as the code to reproduce all of our experimental results are available at https://github.com/aemami1/KnowRef.} We designed this task to be challenging, large-scale, and based on natural text.
The main contributions of this paper are as follows:
\begin{enumerate}
\item We develop mechanisms by which we construct a human-labeled corpus of  8,724 Winograd-like text samples whose resolution requires significant common sense and background knowledge. As an example:

\emph{Marcus is undoubtedly faster than Jarrett right now but in [his] prime the gap wasn't all that big.} (answer: Jarrett)
%Evidenced by the example above, these samples require significant common sense and background knowledge to solve. We generate samples semi-automatically (which makes the dataset easily scalable), automatically scraping candidate sentences from a single corpus, automatically altering the antecedents in the sentences to be identical as far as gender, number, and semantic class, and validating the resulting sentences manually (if they are easy to resolve with common sense) using annotators. 
%\item We demonstrate the difficulty of our task through the performance of six state-of-the-art methods. These fall significantly below human performance.
\item We propose a task-specific metric called \emph{consistency} that measures the extent to which a model uses the full context (as opposed to a surface cue) to make a coreference decision. We use this metric to analyze the behavior of state-of-the-art methods and demonstrate that they generally under-utilize context information.
%\item We propose a novel data-augmentation technique inspired by \citet{trichelair2018evaluation}, called ``antecedent switching,'' to expand our corpus and to further deter models from exploiting surface cues.
\item We find that a fine-tuned version of the recent large-scale language model, BERT \cite{devlin2018bert}, performs significantly better than other methods on \corpusname{}, although with substantial room for improvement to match human performance.
\item We demonstrate the benefits of a data-augmentation technique called \emph{antecedent switching} in expanding our corpus, further deterring models from exploiting surface cues, as well as in transferring to models trained on other co-reference tasks like GAP, leading to state-of-the-art results.
%\item We show that transferring models trained on our corpus to other co-reference tasks like GAP yields state-of-the-art results, and that the benefits of our data-augmentation technique also transfers.

\end{enumerate}

\section{Related Work}

\subsection{General coreference resolution}

Automated techniques for standard coreference resolution --- that is, the task of correctly partitioning the entities and events that occur in a document into resolution classes --- date back to decision trees and hand-written rules \cite{hobbs1977pronoun,mccarthy1995using}.  The earliest evaluation corpora were the Message Understanding Conferences (MUC) \cite{grishman1996message} and the ACE \cite{doddington2004automatic}. These focused on noun phrases tagged with coreference information, but were limited in either size or annotation coverage. 

The datasets of \citet{pradhan2011conll,pradhan2012conll} from the CoNLL-2011 and CoNLL-2012 Shared Tasks were proposed as large-scale corpora with high inter-annotator agreement.
They were constructed by restricting the data to coreference phenomena with highly consistent annotations, and were packaged with a standard evaluation framework to facilitate performance comparisons. 

The quality of these tasks led to their widespread use and the emergence of many resolution systems, ranging from hand-engineered methods to deep-learning approaches. The multi-pass sieve system of \citet{raghunathan2010multi} is fully deterministic and makes use of mention attributes like gender and number; it maintained the best results on the CoNLL 2011 task for a number of years \cite{lee2011stanford}.
Later, lexical learning approaches emerged as the new state of the art \cite{durrett2013easy}, followed more recently by neural models \cite{wiseman2016learning,clark2016deep}. The current state-of-the-art result on the CoNLL 2012 task is by an end-to-end neural model from \citet{lee2018higher} that does not rely on a syntactic parser or a hand-engineered mention detector.

\subsection{Gender bias in general coreference resolution}

\citet{zhao2018gender} observed that state-of-the-art methods for
coreference resolution become gender-biased, exploiting various stereotypes that leak from society into data. %They test two hypotheses for the source of this bias: the training set and external information sources (e.g., word embeddings).
They devise a dataset of 3,160 manually written sentences called \textit{WinoBias} that serves both as a gender-bias test for coreference resolution models and as a training set to counter stereotypes in existing corpora (i.e., the two CoNLL tasks). %This new dataset contains 3160 sentences written manually for the task and covers cases that require an understanding of both semantics and syntax. 
%The sentences contain entities corresponding to people referred to by occupation
The following example is representative:
\begin{exe}
\ex The physician hired the secretary because
\underline{he} was overwhelmed with clients.
\ex The physician hired the secretary because
\underline{she} was overwhelmed with clients.
\end{exe}
Experiments conducted on various models demonstrated that an end-to-end neural model \cite{lee2017end} maintains its performance without the gender bias when trained partially on both the previous datasets and on \textit{WinoBias}.

%The goal is to obtain a balanced dataset without gender stereotypes. Experiments were conducted using three different models: the Stanford coreference resolution system \cite{raghunathan2010multi}, the system of \citet{durrett2013easy}, and the recent end-to-end system \cite{lee2018higher}.

A concurrent work by \citet{rudinger2018gender} also proposed an empirical study of the biases in coreference resolution systems. In contrast to \citet{zhao2018gender}, who attribute the bias in part to the datasets, they conjecture that the gender bias comes primarily from the models themselves. Based on statistics from the Bureau of Labor, they show that various systems all exhibit significant gender bias. %They study three different methods: the rule-based Stanford coreference model \cite{raghunathan2010multi}, statistical methods \cite{bjorkelund2014learning,durrett2013easy} with hand-crafted features, and a neural model \cite{clark2016deep}. 

This work on gender stereotypes provides some insight into the behavior of current models. In the example above, if \textit{she} is predicted incorrectly to refer to \textit{the secretary}, it is likely because the model learned a representation for the secretary profession that encodes gender information. Current models do not capture the context nor the relation between \textit{was overwhelmed}  and \textit{hired} that lead to the correct resolution. The subject of our work is to investigate the potential for models to capture contextual relationships instead of cues from, e.g., gender stereotypes. Unlike WinoBias, our task is composed of passages that occur naturally in text and it is several times larger.
%Moreover, we devise an experiment to show empirically that current state-of-the-art models often rely on the entity itself and ignore the context when making a decision.
%, which can partially explain the results they obtain.

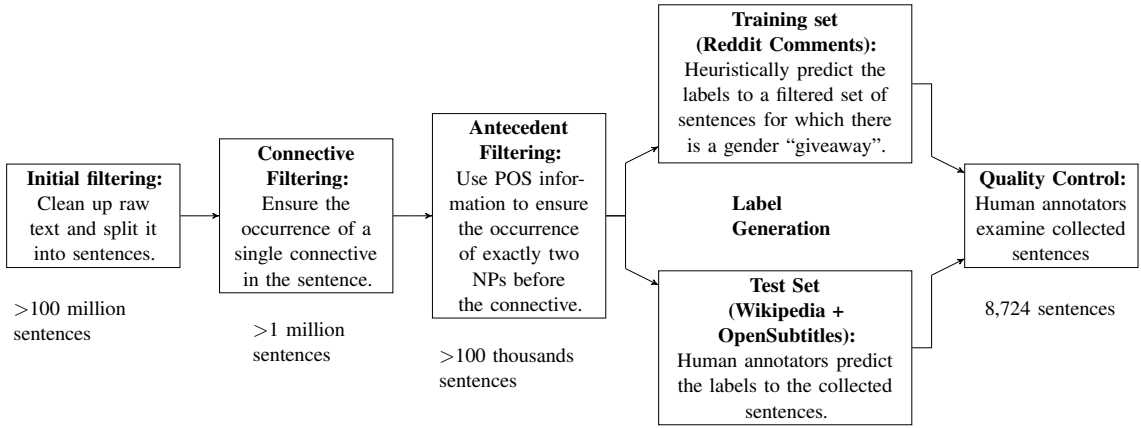
\begin {figure*}[!ht]
\centering
\scalebox{0.70}{
 \begin{tikzpicture}
    [
    mynode/.style={rectangle, draw, align=center, text width =3cm ,minimum width=3cm, minimum height=1cm},
    widenode/.style={rectangle, draw, align=center, text width =4.5cm ,minimum width=4.5cm, minimum height=1cm}
]
    \node[mynode] (a) at ($(0, 0)$) {\textbf{Initial filtering:}\\ Clean up raw text and split it into sentences.};
    \node[mynode] (b) at ($(a) + (4, 0)$) {\textbf{Connective Filtering:}\\ Ensure the occurrence of a single connective in the sentence.};
    \node[mynode] (c) at ($(b) + (4, 0)$) {\textbf{Antecedent Filtering:}\\ Use POS information to ensure the occurrence of exactly two NPs before the connective.};
    \node[widenode] (d1) at ($(c) + (5, 2.5)$) {\textbf{Training set\\(Reddit Comments):}\\Heuristically predict the labels to a filtered set of sentences for which there is a gender ``giveaway".};
    \node[widenode] (d2) at ($(c) + (5, -2.5)$) {\textbf{Test Set \\(Wikipedia + OpenSubtitles):}\\
Human annotators predict the labels to the collected \\sentences.};
    \node[mynode] (e) at ($(c) + (10, 0)$) {\textbf{Quality Control:}\\ Human annotators examine collected sentences};

    \draw (a) -> ($(a)!0.5!(b)$) edge[->] (b);
    \draw (b) -> ($(b)!0.5!(c)$) edge[->] (c);
    \draw (c) -| ($(c)!0.4!(d1)$) edge[->] (d1);
    \draw (c) -| ($(c)!0.4!(d2)$) edge[->] (d2);
    \draw (d1) -| ($(d1)!0.55!(e)$) edge[->] (e);
    \draw (d2) -| ($(d2)!0.55!(e)$) edge[->] (e);

    \node[below =5mm of a,text width=3cm,fill=white]
  (why1) {$>$100 million \\sentences};
  \node[below=5mm of b,text width=2cm,inner sep=.05cm,fill=white]
  (why2) {$>$1 million sentences};
  \node[below=5mm of c,text width=3cm,inner sep=.05cm,fill=white]
  (why3) {$>$100 thousands sentences};
  \node[below=5mm of e,text width=2.5cm,fill=white]
  (why4) {8,724 sentences};
  \node[below=0.5cm of d1,text width=2cm]
  (why4) {\textbf{Label \\Generation}};

\end{tikzpicture}
}
\caption{The corpus construction process for \corpusname{}}
\label{fig:corpcons}
\end{figure*}

\subsection{Difficult cases in coreference resolution}

As the creators of the CoNLL tasks note, most coreference techniques rely primarily on surface-level features, like the proximity between mentions, or shallow semantic features like number, gender, named entities, semantic class, etc., rather than knowledge and context.
%They posit that closing the knowledge gap is essential for the advancement of coreference resolution systems, and that an important step forward includes their large, standardized corpora \cite{pradhan2011conll}.

To address this, \citet{levesque2011winograd} manually constructed a dataset of challenging pronoun disambiguation problems called the Winograd Schema Challenge. The goal was that any successful system would necessarily use common-sense knowledge. Although the WSC is an important step in evaluating systems en route to human-like language understanding, its size and other characteristics are a bottleneck for progress in pronoun disambiguation \cite{trichelair2018evaluation}. A Winograd-like expanded corpus was proposed by \citet{rahman2012resolving} to address the WSC's size limitations; however, systems that perform well on the expanded dataset do not transfer successfully to the original WSC \cite{rahman2012resolving,peng2015solving}, likely due to loosened constraints in the former.\par 
The task that we propose distinguishes itself from the WSC by building on sentences that occur in natural text. This yields highly diverse problem instances. It is particularly important that, as well as being challenging, tasks are representative of natural text, so that improvements are more likely to transfer to the full coreference setting. %The process by which we construct our corpus is also largely automatic, so it is amenable to significant scaling.

Recently, \citet{webster2018mind} presented a corpus called GAP that consists of 4,454\footnote{In GAP, one unique coreference instance corresponds to two pronoun-name pairs, for which they report 8,908 pairs.} unique binary coreference instances from English Wikipedia.
It is meant to address gender bias and the described size limitation of the WSC.
For instance, it exposes the unbalanced performance of current state-of-the-art resolvers, which more accurately resolve masculine pronouns than feminine pronouns.
%Three annotators were used to label GAP sentences. In 73\% of cases there was full agreement, but in 25\% of cases only two annotators agreed.
%In addition, GAP is not entirely a binary-choice coreference resolution task akin to the WSC, since the corpus contains instances where neither of the two entities co-refer with the ambiguous pronoun.
As for the difficulty of the task, the models tested on GAP were not trained directly on the corpus, which does not give a clear picture of the task's difficulty. A simple heuristic called \textit{Parallelism+URL}, which is based on using the syntactic distance between antecedents and the target pronoun, is so far the strongest GAP baseline, at above 70\% accuracy. This suggests that GAP is vulnerable to exploits that circumvent a need for knowledge, albeit not the gender and number cues that coreference resolvers have exploited before. Finally, our corpus construction process differs from that of GAP's by more strictly requiring that the sentences are in WSC-format, that is, there are exactly two named entities that occur strictly before the pronoun and only one of which may co-refer with the pronoun (in GAP, the pronoun may occur between and before the named entities and may in fact co-refer with both named entities). In addition, our corpus construction process exploits the fact that the named entities can be replaced with any name in order to increase the task difficulty by automatically removing gender giveaways as well as to significantly increase the size of the corpus by switching the named entities to create a new task instance.

As such, our paper seeks to explore a wider problem of which gender bias may be one facet: current models do not effectively abstract away from the entities (and instead rely on exploits using gender or plurality) to make the coreference resolution. By developing a benchmark task consisting strictly of sentences for which such cues are ineffective, we seek to challenge and potentially improve current coreference resolution models. In addition, based on our new benchmark, \corpusname{}, we introduce a data-augmentation mechanism, called \textit{antecedent switching}, to encourage models to perform this abstraction. %We demonstrate that training a model using \corpusname{} and the introduced data-augmentation trick not only improves general performance but also mitigates the gender bias issue.
%Our \corpusname{} benchmark is distinct from GAP in several ways. Most importantly, the focus of our corpus is not to demonstrate or correct for gender bias as it is to serve as the first demonstrably difficult and large-scale common-sense reasoning task that is robust to exploitations on surface cues such as the gender, number and the semantic class of competing antecedents. In addition, our corpus-construction procedure gives rise to a core measure we call \textit{consistency} that can be used alongside accuracy to evaluate performant systems. Particularly, it allows for a deeper insight to the extent in which a system incorporates the full context (as opposed to a surface cue) to solve the problem instance and in doing so, be \textit{consistent} in the way that it resolves it.

\section{The \corpusname{} Coreference Task}

We develop a coreference task called \corpusname{} that features 8,724 difficult pronoun disambiguation problems.  Each instance is a short passage containing a target pronoun that must be correctly resolved to one of two possible antecedents.

Formally, each problem instance can be described as a tuple $P=\{S,C_1,C_2,T,K\}$, where $S$ is the sentence, $C_1$ and $C_2$ are the candidate antecedents, $T$ is the target pronoun to be resolved to one of $C_1$ and $C_2$, and $K$ indicates the correct antecedent. Note that $C_1$, $C_2$, $T$ and $K$ appear in $S$.
\corpusname{} provides $\{S,C_1,C_2,T\}$ as input for models, which must predict $K$ (e.g., as the output of a binary classification over $C_1,C_2$).
A representative sentence $S$ is the following.
\begin{exe}
    \ex \label{ex-1} \{Paul\} helped \{Lionel\} hide when [he] was pursued by the authorities.
\end{exe}
Here, $C_1=\text{Paul}$, $C_2=\text{Lionel}$, $T=\text{he}$, and $K=C_2=\text{Lionel}$.

We control the text so as not to give away the pronoun's correct antecedent in surface-level cues involving syntactic salience or the number and gender of the antecedent. Successful systems must instead make use of the context, which may require world knowledge and common-sense inferences; i.e., that someone who is being helped to hide may be one who is being pursued by the authorities.

In the following section, we describe the methodology used to construct our corpus, provide a glimpse of a few of its instances and their resolution rationales, outline the task's evaluation criteria, and describe its characteristics.

\begin{table*}[ht]
\small 
\begin{center}
\begin{tabu}to\linewidth{@{}X[l]X[l,4]@{}} 
\toprule

\corpusname{} Example 1: & \{Radu\} appeared to be killed by \{Brother Paulo\}, but [he] reappears a short while later injured, but alive. ($K=\text{Radu}$) \\
Original sentence: & Radu appeared to be killed by Sister Paula, but he reappears a short while later injured, but alive.\\

\midrule

\corpusname{} Example 2: & \{Wanda\} tries to apologize to \{Rose\}, but [she] refuses to accept.  ($K=\text{Rose}$) \\
Original sentence: & Warren tries to apologize to Rose, but she refuses to accept.\\

\midrule

\corpusname{} Example 3: & \{Tom\} arrives to where \{Alex\} was tied, but [he] has come free of his lead. ($K=\text{Alex}$)\\
Original sentence: & Tom arrives to where Vanessa was tied, but she has come free of her lead.\\

%\midrule

%\corpusname{} Example 4: & \underline{Carl} then realizes that \underline{Peter} is innocent if [he] genuinely believed him to be guilty. (Answer: Peter)\\
%Original sentence: & Carla then realizes that Peter is innocent if he genuinely believed her to be guilty.\\

\bottomrule
\end{tabu}
\caption{Examples of \corpusname{} instances. %The target pronouns are boldfaced and the special words underlined.
}
\label{tab:examples}
\end{center}
\vskip -.1in
\end{table*}

\subsection{Corpus construction}

To construct \corpusname{}, we scrape text samples from a large collection of documents: the combination of 2018 English Wikipedia, OpenSubtitles, and Reddit comments dating from 2006--2018. We filter this text through a multi-stage process to ensure quality and diversity as depicted in Figure \ref{fig:corpcons}, and described in more detail below.

%We remove markup, non-ASCII characters, parenthetical expressions, headings, and lists, and split the text into sentences.  We test that each sentence contains at least two noun-phrase antecedents and a pronoun (as in the example in the previous section). For these sentences,  we use a combination of automatic labelling heuristics and human annotation to determine the antecedent that correctly refers to the pronoun.

\subsubsection{Initial Filtering}
After removing markup, non-ASCII characters, parenthetical expressions, headings and lists, we split the text into sentences. We keep sentences of token length between 9 and 33 words after na\"ive tokenization, which start with an upper case letter, and which contain no math. 

\subsubsection{Connective Filtering}
Our first substantial filtering step uses regular expressions to ensure that each passed sentence contains connectives.\footnote{comma, semicolon, \textit{or}, \textit{since}, \textit{but}, \textit{because}, \textit{although}, etc.} We use a regular expression to ensure that there is only one connective cluster (e.g. ``, and though''), and that there are at least two non-stopwords before this connective and a pronoun after it. As a final check, we ensure that no pronoun occurs before the connective, which tends to remove sentences which are not self-contained.

\subsubsection{Antecedent Filtering}
On the remaining set of sentences, we use Stanford's Maxent tagger \cite{toutanova2003feature} to infer a flat part-of-speech (POS) labelling. Using the inferred POS tags, we ensure that there are exactly two noun phrases (NPs) before the connective that do not re-occur after it (a re-occurrence after the connective means that the pronoun likely refers to the non-repeated noun phrase). %Next, we heuristically determine the number, type, and gender of the noun phrases. We ensure that we can use the pronoun to distinguish between the two noun phrases (e.g. ``they'' indicates plural, ``he/him'' indicates male person).

The mentioned checks resulted in roughly 100,000 sentences across all three corpora. At least some of these remaining sentences have similar properties to Winograd schema sentences; that is, the two noun phrases (NPs) and the pronoun share the same type. From here, we keep only sentences where the type indicates that both NPs correspond to persons, which further filters the remaining sentences. We do this because NPs that denote people are often named entities or can easily be replaced by named entities without loss of information. We targeted these instances also because we investigate how resolution systems use gender cues and most gendered pronouns occur with person-type NPs.

\subsubsection{Label Generation}
We generate our training and test sets from distinct sources of text using two different methods. 

\paragraph{Training set:}We automatically collect 70,000 sentences from Reddit that have passed the filters described above, and filter these down to roughly 7,500 sentences for which the antecedents are named entities of different genders. We use a Python library\footnote{\url{https://pypi.org/project/SexMachine/}} to infer the genders, based on a list of 40,000 names categorized as female or male compiled by Jörg Michael. Given the pronoun and the distinct predicted genders for the antecedents, we can infer the label for the pronoun's correct resolution with high accuracy and without the need for expensive human annotation. After assigning this label, we remove the gender giveaway by replacing one of the named entities so that both entities and the pronoun all match in gender (e.g., in a sentence with ``James'', ``Jessica'', and ``she'' as the NPs and pronoun, we replace ``James'' with ``Jane''). These sentences form our training set. To assess its quality, we gave an annotator a random sample of 100 training instances with their heuristically determined labels. The annotator then evaluated each sentence as ``correctly labelled", ``incorrectly labelled", or ``unresolvable" if neither of the two candidates were more suitable than the other to corefer with the pronoun.\footnote{The details and result of this quality-testing study will also be made public along with the code and dataset.} In total, 86\% of the instances were deemed to be labelled correctly, 11\%  incorrectly labelled, and 3\%  were not resolvable, implying that our automatic selection heuristic is strong but imperfect.

\paragraph{Test set:}Human annotators examined all collected sentences for quality control. We also use a source for the test sentences that is distinct from that of the training set, directing our pipeline to collect sentences from Wikipedia and OpenSubtitles rather than Reddit. This is to ensure that stylistic cues common in the training source cannot be exploited by models at test time. In total, roughly 10,000 candidate sentences were extracted initially. As before, we automatically remove gender giveaways by replacing the named entities with names of the same gender, rendering the pronoun ambiguous. Then, six human annotators predicted which antecedent was the correct coreferent of the pronoun for a sample of 2,000 candidate sentences, or they labeled the sentence with ``neither" (in the case where neither antecedent feasibly corefers with the pronoun) or ``unclear" (if the sentence was not intelligible). Sentences that have a strong agreement from 5 or more annotators on a single antecedent (and which are not labeled as ``neither" or ``unclear") are kept for testing. This yielded 1,269 test sentences. We measured high inter-annotator agreement on the test set with a Fleiss’ Kappa of $\kappa=0.78$.

\begin{table}[t]
\begin{center}
\begin{tabu}to\linewidth{@{}X[3,l]X[2,r]@{}}
\toprule
Sentence Characteristic            &  \% of Data    \\ \midrule
Masculine target pronouns               & 52.7        \\
Feminine target pronouns               & 47.3   \\     \cmidrule(r){1-1}
First Antecedent Correct & 50.7         \\
Second Antecedent Correct             & 49.2         \\ \bottomrule
\end{tabu}
\caption{Characteristics of the dataset, in terms of pronoun distribution and correct label.}
\label{tab:characteristics}
\end{center}

\end{table}

Our pipeline thus yields a total of 8,724 sentences (7,455 training and 1,269 test) whose pronoun disambiguation should not be clear from shallow features like gender, number, and semantic type -- they should instead require varying degrees of external knowledge. These sentences constitute the \corpusname{} corpus. Examples of some instances are given in Table~\ref{tab:examples}. As these examples reveal, each instance may require a unique bit of common sense knowledge to resolve.

In the first example, the common understanding that death (by way of killing) causes a disappearance helps us to conclude that Radu, the victim of murder, is the one to who reappears.

In the next example, human readers recognize that to \textit{accept} is something one does with an \textit{apology}. Therefore, \textit{she} refers to the one that accepts the apology, i.e., Rose.

For the third example, an understanding that \textit{being tied} is related to being deprived of freedom leads us to conclude that Alex has come free.

%In the fourth example, we first see that Peter is realized to be innocent by Carl and that the second clause is there to justify the realization. In reading the second clause, a candidate person (either Peter or Carl) is said to have genuinely believed the other to be guilty. An important trait of an innocent person can be their genuine belief that they are indeed innocent, which, in turn, may also result in their genuine belief that someone else is guilty. Accordingly, it fits better that it was Peter who had this genuine belief, and this is what gives Carl the justification he needs to realize Peter's innocence. 

\subsection{Task Characteristics}

\begin{table*}[t]
\begin{center}
\begin{tabu}to \linewidth{@{}X[l,3]*5{X[1.5,c]}@{}}
\toprule
Model                                &  Both Antecedents Predicted  & No Decision    & Incorrect Decision & Correct Decision &  Task-\linebreak Specific \linebreak Accuracy  \\ \midrule
Random          & --          & -- & -- & -- & 0.50  \\
Human\footnotemark[5]             & --          & -- & -- & -- & 0.92 \\
\cmidrule(r){1-1}
Rule                     & 0.001 & 0.12 & 0.43  & 0.45 & 0.52 \\ 
Stat               & 0.006          & 0.09 & 0.45 & 0.45 & 0.50  \\ 
Deep-RL & 0.001         & 0.09 & 0.46 & 0.45 & 0.49  \\
Latent & 0.000         & 0.12 & 0.41 & 0.47 & 0.54  \\
E2E (CoNLL only) & 0.01         & 0.42 & 0.23  & 0.35 & 0.60  \\
\cmidrule(r){1-1}
E2E (\corpusname{}) & 0.000         & 0.26 & 0.31  & 0.43 & 0.58 \\
E2E (\corpusname{}+CoNLL) & 0.000     & 0.19 & \textbf{0.28}  & 0.52 & \textbf{0.65}  \\
BERT (\corpusname{}) & 0.000         & 0.000 & 0.39  & \textbf{0.61} & 0.61  \\
 \bottomrule
\end{tabu}
\caption{Coverage and performance of various representative systems on the \corpusname{} Test set.}
\label{tab:performance}
\end{center}

\end{table*}

In Table \ref{tab:characteristics}, we report several statistical characteristics of the data. These suggest a near-equal distribution of feminine and masculine target pronouns (\textit{he/him/his} vs. \textit{she/her}) as well as an equal distribution of the two labels, which keeps chance-based performance at 50\% expected accuracy.

\subsection{Evaluation}
Our task requires a model to choose between two candidates, but classical coreference models build clusters of expressions that refer to the same entity. With respect to our setting, several errors can be made by these existing models: predicting that the two entities and the pronoun share a similar cluster (\textit{Both Antecedents Predicted}), that none of the two candidates shares a cluster with the pronoun (\textit{No Decision}), or creating a cluster that contains the pronoun with the wrong candidate (\textit{Incorrect Decision}). To obtain a score specific to our task, we compute a \textit{Task-Specific Accuracy} which discards all of the cases in which the model makes no decision relevant to the target pronoun or chooses both entities as co-referring to the target pronoun.

%Given that there may exist a degree of subjectivity to the answers, we also establish human baseline performance on the task. To do so, we randomly sample 100 examples from the test set and ask six native English speakers to choose the right antecedent. The answer was considered correct if the label agrees with the strong majority (5/6 annotators). This results in a human performance of 92\% as reported in table \ref{tab:performance}.

\section{Experiments and Results}

In this section, we compare the performance of five representative coreference systems on our task: Stanford’s rule-based system \cite{raghunathan2010multi} (\textbf{Rule}), Stanford’s statistical system \cite{clark2015entity} (\textbf{Stat}), \citet{clark2016deep}’s deep reinforcement learning system (\textbf{Deep-RL}), \citet{martschat2015latent}’s latent tree model (\textbf{Latent}), and \citet{lee2018higher}’s end-to-end neural system (\textbf{E2E}). We also report the accuracy of the state-of-the-art model, \textbf{E2E}, after retraining on \corpusname{} and on \corpusname{}+CoNLL.

Additionally, we develop a task-specific model for \corpusname{}: a discriminatively trained fine-tuned instance of Bidirectional Encoder Representations from
Transformers (\textbf{BERT}) \cite{devlin2018bert}. We train our task-specific \textbf{BERT} according to recent work on language models (LMs) for the WSC \cite{trinh2018simple}. We first construct a modified version of the data wherein we duplicate each sentence, replacing the pronoun with one of the two antecedents in each copy. The task, akin to NLI, is then to predict which of the two modified sentences is most probable. To compute probabilities, we add a softmax layer with task-specific parameter vector $v\in\mathcal{R}^H$. Denote by $h_{S1}\in\mathcal{R}^H$ (respectively $h_{S2}$) the  final hidden state for the sentence copy with the pronoun replaced by the first antecedent (respectively the second). Then the probability assigned to the first antecedent is
\begin{align}
    P_1 = \frac{e^{v^\top h_{S1}}}{e^{v^\top h_{S2}}+e^{v^\top h_{S2}}}.
\end{align}
The probability assigned to the second antecedent is $P_2=1-P_1$. We use $H=768$ hidden units in our BERT implementation and learn $v$ by minimizing the binary cross entropy with the ground-truth antecedent labels (in one-hot format).

\paragraph{Human Performance:} We determined human performance on \corpusname{} by collecting the predictions of six native English speakers on a randomly generated sub-sample of 100 problem instances; we consider correct those predictions that agreed with the majority decision and matched the ground-truth label derived from the original sentence. 
We report the performance of the five coreference systems and the human baseline in Table~\ref{tab:performance}.\par The human performance of 0.92 attests to the task's viability. The performance of the automatic systems pretrained on CoNLL, at random or slightly above random, demonstrates that state-of-the-art coreference resolution systems are unable to solve the task. This suggests the existence in the wild of difficult but realistic coreference problems that may be under-represented in CoNLL.\par
After training on \corpusname{}, \textbf{E2E} improves by more than 5\% in task-specific accuracy. We can infer from this result that the model can make some use of context to make predictions if trained appropriately, but that the CoNLL shared tasks may not contain enough of such instances for models to generalize from them. Finally our task-specific model reaches an accuracy of at best 65\%, far below human performance despite having access to the two candidates.
\footnotetext[5]{This is an estimate based on a subsample of the data.}

\subsection{Analysis by Switching Entities}
Inspired by \citet{trichelair2018evaluation}, we propose to use a task-specific metric, \textit{consistency}, to measure the ability of a model to use context in its coreference prediction, as opposed to relying on gender and number cues related to the entities. Accounting for this is critical, as we desire models that can capture social, situational, or physical awareness.\par
To measure consistency in the \corpusname{} corpus, we duplicate the data set but switch the candidate antecedents each time they appear in a sentence. This changes the correct resolution. If a coreference model relies on knowledge and contextual understanding, its prediction should change as well, thus it could be called \textit{consistent} in its decision process. If, however, its decision is influenced solely by the antecedent, its output would stay the same despite the change in context induced by switching.
We define the \textit{consistency} score as the percentage of predictions
that change from the original instances to the switched instances.
%We hypothesize that the current systems mostly rely on gender and number cues that are related to the entities themselves and not the context. This causes performance to suffer when sentences contain gender-unbiased entities. We propose an experiment to validate our hypothesis. We use the \corpusname{} corpus and switch the two candidates every time they appear in a sentence. This changes the correct resolution; if a coreference model relies on knowledge and contextual understanding, its prediction should change as well, and it could be said to be``consistent" in its decision process, while if it relies on entity information, its output should stay the same. We thereby propose a new metric called ``consistency" which measures the extent to which a model relies on the full context of sentences in a task to make a decision.  \\ 
An example of a switching is:
\begin{exe}
    \ex \label{ex-2} 
    \textbf{Original}: \{Alex\} tells \{Paulo\}, but [he] does not believe him.\\
    \textbf{Switched}: \{Paulo\} tells \{Alex\}, but [he] does not believe him.
\end{exe}
The correct answer switches from $K=\text{Paulo}$ to $K=\text{Alex}$.

\begin{table}[t]
\begin{center}
\begin{tabu}to\linewidth{@{}X[l,3]X[c,1]@{}}
\toprule
Model                                   &   Consistency  \\ \midrule
Rule            & 0\%  \\
Stat               & 76\%  \\
Deep-RL & 66\%  \\
Latent & 78\% \\
E2E & 62\%\\
\cmidrule(r){1-1}
E2E (\corpusname{}) & 66\%  \\
E2E (\corpusname{}+CoNLL) & 67\%  \\
BERT (\corpusname{}) & 69\%  \\
\bottomrule
\end{tabu}
\caption{The sensitivity of various systems to the instance antecedents, according to the number of changed decisions when the antecedents are switched. Higher is better.}
\label{tab:sensitivity}
\end{center}
\end{table}

Table~\ref{tab:sensitivity} shows the consistency scores of the various baseline models evaluated on the original and switched duplicates of \corpusname{}. The rule-based system \cite{raghunathan2010multi} always resolves to the same entity, suggesting that context is ignored. Indeed, the mechanisms underlying this model mostly rely on a gender and number dictionary \cite{bergsma2006bootstrapping}. This dictionary informs a count-based approach that assigns a masculine, feminine, neutral, and plural score to each word. If the pronoun is \textit{his}, the candidate with the higher masculine score is likely to be linked to the pronoun.\par

The other models, Stat, Deep-RL, E2E, Latent and BERT are much more robust to the switching procedure, demonstrating that the resolution partially relies on context cues.  Regarding \textbf{E2E}, we can observe that training the model on \corpusname{} forces the model to rely more on the context, leading to an improvement of 5\%. It further demonstrates the usefulness of the corpus to obtain a better representation of the context.
%Yet, it is surprising that the model proposed by \citet{martschat2015latent}, which relies only on the previous and next token of each candidate as a context feature, outperforms the end-to-end model. The latter uses a Bidirectional LSTM to build a context-dependent representation of each candidate. It is likely that the model adapts to its training set, OntoNotes 5.0, by focusing on gender/number cues that underpin the majority of pronoun disambiguation.

\subsection{Data Augmentation by Switching}
Inspired by the switching experiment, we propose to extend the \corpusname{} training set by switching every entity pair (thereby doubling the number of instances). We hypothesize that this data augmentation trick could force the model to abstract away from the entities to the context in order to boost performance, since it encounters the same contextual scenario in the doubled sentences.
%It is also a way to test the ability of a model to capture knowledge and transferring it when facing similar concept.\\

\begin{table}[t]
\begin{center}
\begin{tabu}to\linewidth{@{}X[l,3]X[c,1.8]X[c,0.4]X[c,2]@{}}
\toprule
Model                                   &   Accuracy & $\Delta$ & Consistency\\ \midrule
BERT (\corpusname{})  & 71\% & +10\% & 89\% \\
E2E (\corpusname{}) & 61\% & +3\% & 71\% \\
E2E (\corpusname{}+CoNLL) & 66\% & +1\% & 75\% \\

\bottomrule
\end{tabu}
\caption{Accuracy on the \corpusname{} test set for each model after augmenting the training set, as well as the difference from the result without data augmentation.}
\label{tab:afterswitch}
\end{center}
\end{table}

Training on the augmented data, we observe an improvement of 10\% for fine-tuned BERT (Table~\ref{tab:afterswitch}), yielding a task-specific accuracy of 71\% on the \corpusname{} test set. The improvement in accuracy is marginal for \textbf{E2E}, but we observe a large gain in consistency.
%demonstrating the ability of the model to rely on the context to choose the right antecedent.
We suspected that the data augmentation trick might also be useful in mitigating a model's gender bias, by encouraging the model to rely more on the context than on gendered entity names. To test this hypothesis, we train the same model with and without the data augmentation trick on the recently released GAP corpus \cite{webster2018mind}.

\begin{table}[ht]
\begin{center}
\begin{tabu}to\linewidth{@{}X[l,5]X[c,1.5]X[c,1.5]@{}}
\toprule
Model &  $\frac{F_1^{F}}{F_1^{M}}$  & $F_1$   \\ \midrule
Parallelism\footnotemark[6] & 0.93 & 66.9 \\
Parallelism+URL\footnotemark[6]  & 0.95 & 70.6  \\
BERT (GAP) & 1.02 & 69.2 \\
BERT (GAP) + Data Aug.  & \textbf{1.00} & \textbf{71.1} \\
\bottomrule
\end{tabu}
\caption{Performance on the GAP test set}
\label{tab:afterswitch-gap}
\end{center}
\end{table}
\footnotetext[6]{Scores reported in the original paper \cite{webster2018mind}}

\begin{table*}[!ht]
\small 
\centering
%\begin{tabu*}{@{}\linewidth}{p{0.12\textwidth}p{0.65\textwidth}p{0.1\textwidth}@{}}
\begin{tabu*}to\linewidth{@{}X[0.15]X[0.65]X[0.2]@{}}
\toprule

Sentence Type & Sentence & Answer \\ \midrule

Original  \newline Switched              & 
{Kara} is in love with {Tanya} but she is too shy to tell [her]. \newline
{Tanya} is in love with {Kara} but she is too shy to tell [her].
 & Tanya \checkmark \newline  Kara \checkmark \newline (consistently correct)
\\ \midrule
Original  \newline Switched              & 
{Peter} had not realised how old {Henry} was until [he] sees his daughter.  \newline
{Henry} had not realised how old {Peter} was until [he] sees his daughter.
 & Henry \xmark \newline  Peter \xmark \newline (consistently incorrect) \\
 \midrule
 Original  \newline Switched              & 
{Poulidor} was no match for {Merckx}, although [he] offered much resistance .
 \newline
{Merckx} was no match for {Poulidor}, although [he] offered much resistance .

 & Poulidor \checkmark \newline  Poulidor \xmark \newline (inconsistent) \\

\bottomrule
\end{tabu*}
\caption{Examples of various success/failure cases of BERT on the \corpusname{} test set}
\label{tab:examples2}
\end{table*}

BERT fine-tuned on GAP achieves a state of the art $F_1$ of 71.1 after data augmentation (Table~\ref{tab:afterswitch-gap}). Not only does the augmentation improve the overall performance (+1.9) but it further balances the predictions' female:male ratio to 1:1.

\subsection{Error Analysis}

We show examples of BERT’s performance (trained on \corpusname{}) on our test set in Table \ref{tab:examples2}. This includes instances on which it succeeds and fails for both original and switched sentences. 
In general, it is not clear why certain instances are more difficult for BERT to resolve, although training BERT on the augmented, switched corpus significantly reduces the frequency of inconsistent resolutions (from 31\% to 11\%).

These examples illustrate how challenging certain real-world situations can be for models to understand, compared to humans who can reason about them with ease.

\section{Conclusion}
We present a new corpus and task, \corpusname{}, for coreference resolution. Our corpus contains difficult problem instances that require a significant degree of common sense and world knowledge for accurate coreference link prediction, and is larger than previous similar datasets. Using a task-specific metric, consistency, we demonstrate that training coreference models on \corpusname{} improves their ability to build better representations of the context. We also show that progress in this capability is linked to reducing gender bias, with our proposed model setting the state of the art on GAP. \par
In the future, we wish to study the use of \corpusname{} to improve performance on general coreference resolution tasks (e.g., the CoNLL 2012 Shared Tasks). We also plan to develop new models on \corpusname{} and transfer them to difficult common sense reasoning tasks. %We presented a coreference task that features high human performance and inter-annotator agreement but random or slightly above-random performance for various state-of-the-art resolution systems, including rule-based, graphical, and neural models. In addition, we demonstrated that the antecedents of a sentence seem to have the greatest impact on the resolution decision of these models, a behaviour that suggests their reliance on shallow semantic features like number and gender. In turn, models that exploit these features may perform very well on certain coreference tasks, but, not on those for which a significant degree of common sense and world knowledge is required.
%In the future, we plan to develop a larger corpus of at least 100,000 problem instances, by using additional text-data sources (besides Wikipedia, OpenSubtitles and Reddit). This would represent a corpus even more amenable to powerful neural models, which could serve as an important training set for models that have thus far performed poorly on difficult coreference problems like the Winograd Schema Challenge. 

\section*{Acknowledgements}
This work was supported by the Natural Sciences and Engineering Research Council of Canada and by Microsoft Research. Jackie Chi Kit Cheung is supported by the Canada CIFAR AI Chair program.

\bibliography{acl2019}

\begin{thebibliography}{33}
\expandafter\ifx\csname natexlab\endcsname\relax\def\natexlab#1{#1}\fi

\bibitem[{Bailey et~al.(2015)Bailey, Harrison, Lierler, Lifschitz, and
  Michael}]{bailey2015winograd}
Dan Bailey, Amelia Harrison, Yuliya Lierler, Vladimir Lifschitz, and Julian
  Michael. 2015.
\newblock The winograd schema challenge and reasoning about correlation.
\newblock In \emph{Working Notes of the Symposium on Logical Formalizations of
  Commonsense Reasoning}.

\bibitem[{Bean and Riloff(2004)}]{bean2004unsupervised}
David Bean and Ellen Riloff. 2004.
\newblock Unsupervised learning of contextual role knowledge for coreference
  resolution.
\newblock In \emph{Proceedings of the Human Language Technology Conference of
  the North American Chapter of the Association for Computational Linguistics:
  HLT-NAACL 2004}.

\bibitem[{Bergsma and Lin(2006)}]{bergsma2006bootstrapping}
Shane Bergsma and Dekang Lin. 2006.
\newblock Bootstrapping path-based pronoun resolution.
\newblock In \emph{Proceedings of the 21st International Conference on
  Computational Linguistics and the 44th annual meeting of the Association for
  Computational Linguistics}.

\bibitem[{Clark and Manning(2015)}]{clark2015entity}
Kevin Clark and Christopher~D Manning. 2015.
\newblock Entity-centric coreference resolution with model stacking.
\newblock In \emph{Proceedings of the 53rd Annual Meeting of the Association
  for Computational Linguistics and the 7th International Joint Conference on
  Natural Language Processing}.

\bibitem[{Clark and Manning(2016)}]{clark2016deep}
Kevin Clark and Christopher~D Manning. 2016.
\newblock Deep reinforcement learning for mention-ranking coreference models.
\newblock In \emph{Proceedings of the 2016 Conference on Empirical Methods in
  Natural Language Processing}.

\bibitem[{Devlin et~al.(2018)Devlin, Chang, Lee, and
  Toutanova}]{devlin2018bert}
Jacob Devlin, Ming-Wei Chang, Kenton Lee, and Kristina Toutanova. 2018.
\newblock Bert: Pre-training of deep bidirectional transformers for language
  understanding.
\newblock \emph{arXiv preprint arXiv:1810.04805}.

\bibitem[{Doddington et~al.(2004)Doddington, Mitchell, Przybocki, Ramshaw,
  Strassel, and Weischedel}]{doddington2004automatic}
George~R Doddington, Alexis Mitchell, Mark~A Przybocki, Lance~A Ramshaw,
  Stephanie Strassel, and Ralph~M Weischedel. 2004.
\newblock The automatic content extraction (ace) program-tasks, data, and
  evaluation.
\newblock In \emph{LREC}.

\bibitem[{Durrett et~al.(2013)Durrett, Hall, and
  Klein}]{durrett2013decentralized}
Greg Durrett, David Hall, and Dan Klein. 2013.
\newblock Decentralized entity-level modeling for coreference resolution.
\newblock In \emph{Proceedings of the 51st Annual Meeting of the Association
  for Computational Linguistics}.

\bibitem[{Durrett and Klein(2013)}]{durrett2013easy}
Greg Durrett and Dan Klein. 2013.
\newblock Easy victories and uphill battles in coreference resolution.
\newblock In \emph{Proceedings of the 2013 Conference on Empirical Methods in
  Natural Language Processing}.

\bibitem[{Grishman and Sundheim(1996)}]{grishman1996message}
Ralph Grishman and Beth Sundheim. 1996.
\newblock Message understanding conference-6: A brief history.
\newblock In \emph{COLING 1996 Volume 1: The 16th International Conference on
  Computational Linguistics}, volume~1.

\bibitem[{Hobbs(1977)}]{hobbs1977pronoun}
Jerry~R Hobbs. 1977.
\newblock Pronoun resolution.
\newblock \emph{ACM SIGART Bulletin}.

\bibitem[{Lee et~al.(2011)Lee, Peirsman, Chang, Chambers, Surdeanu, and
  Jurafsky}]{lee2011stanford}
Heeyoung Lee, Yves Peirsman, Angel Chang, Nathanael Chambers, Mihai Surdeanu,
  and Dan Jurafsky. 2011.
\newblock Stanford's multi-pass sieve coreference resolution system at the
  conll-2011 shared task.
\newblock In \emph{Proceedings of the fifteenth conference on computational
  natural language learning: Shared task}.

\bibitem[{Lee et~al.(2017)Lee, He, Lewis, and Zettlemoyer}]{lee2017end}
Kenton Lee, Luheng He, Mike Lewis, and Luke Zettlemoyer. 2017.
\newblock End-to-end neural coreference resolution.
\newblock In \emph{Proceedings of the 2017 Conference on Empirical Methods in
  Natural Language Processing}.

\bibitem[{Lee et~al.(2018)Lee, He, and Zettlemoyer}]{lee2018higher}
Kenton Lee, Luheng He, and Luke Zettlemoyer. 2018.
\newblock Higher-order coreference resolution with coarse-to-fine inference.
\newblock In \emph{Proceedings of the 2018 Conference of the North American
  Chapter of the Association for Computational Linguistics: Human Language
  Technologies}.

\bibitem[{Levesque et~al.(2011)Levesque, Davis, and
  Morgenstern}]{levesque2011winograd}
Hector~J Levesque, Ernest Davis, and Leora Morgenstern. 2011.
\newblock The winograd schema challenge.
\newblock In \emph{AAAI Spring Symposium: Logical Formalizations of Commonsense
  Reasoning}.

\bibitem[{Liu et~al.(2016)Liu, Jiang, Evdokimov, Ling, Zhu, Wei, and
  Hu}]{liu2016probabilistic}
Quan Liu, Hui Jiang, Andrew Evdokimov, Zhen-Hua Ling, Xiaodan Zhu, Si~Wei, and
  Yu~Hu. 2016.
\newblock Probabilistic reasoning via deep learning: Neural association models.
\newblock \emph{arXiv preprint arXiv:1603.07704}.

\bibitem[{Martschat and Strube(2015)}]{martschat2015latent}
Sebastian Martschat and Michael Strube. 2015.
\newblock Latent structures for coreference resolution.
\newblock \emph{Transactions of the Association of Computational Linguistics}.

\bibitem[{McCallum and Wellner(2005)}]{mccallum2005conditional}
Andrew McCallum and Ben Wellner. 2005.
\newblock Conditional models of identity uncertainty with application to noun
  coreference.
\newblock In \emph{Advances in neural information processing systems}.

\bibitem[{McCarthy(1995)}]{mccarthy1995using}
JF~McCarthy. 1995.
\newblock Using decision trees for coreference resolution.
\newblock In \emph{Proc. 14th International Joint Conf. on Artificial
  Intelligence}.

\bibitem[{Morton(2000)}]{morton2000coreference}
Thomas~S Morton. 2000.
\newblock Coreference for nlp applications.
\newblock In \emph{Proceedings of the 38th Annual Meeting on Association for
  Computational Linguistics}. Association for Computational Linguistics.

\bibitem[{Peng et~al.(2015)Peng, Khashabi, and Roth}]{peng2015solving}
Haoruo Peng, Daniel Khashabi, and Dan Roth. 2015.
\newblock Solving hard coreference problems.
\newblock \emph{Urbana}.

\bibitem[{Pradhan et~al.(2012)Pradhan, Moschitti, Xue, Uryupina, and
  Zhang}]{pradhan2012conll}
Sameer Pradhan, Alessandro Moschitti, Nianwen Xue, Olga Uryupina, and Yuchen
  Zhang. 2012.
\newblock Conll-2012 shared task: Modeling multilingual unrestricted
  coreference in ontonotes.
\newblock In \emph{Joint Conference on EMNLP and CoNLL-Shared Task}.

\bibitem[{Pradhan et~al.(2011)Pradhan, Ramshaw, Marcus, Palmer, Weischedel, and
  Xue}]{pradhan2011conll}
Sameer Pradhan, Lance Ramshaw, Mitchell Marcus, Martha Palmer, Ralph
  Weischedel, and Nianwen Xue. 2011.
\newblock Conll-2011 shared task: Modeling unrestricted coreference in
  ontonotes.
\newblock In \emph{Proceedings of the Fifteenth Conference on Computational
  Natural Language Learning: Shared Task}.

\bibitem[{Raghunathan et~al.(2010)Raghunathan, Lee, Rangarajan, Chambers,
  Surdeanu, Jurafsky, and Manning}]{raghunathan2010multi}
Karthik Raghunathan, Heeyoung Lee, Sudarshan Rangarajan, Nathanael Chambers,
  Mihai Surdeanu, Dan Jurafsky, and Christopher Manning. 2010.
\newblock A multi-pass sieve for coreference resolution.
\newblock In \emph{Proceedings of the 2010 Conference on Empirical Methods in
  Natural Language Processing}. Association for Computational Linguistics.

\bibitem[{Rahman and Ng(2009)}]{rahman2009supervised}
Altaf Rahman and Vincent Ng. 2009.
\newblock Supervised models for coreference resolution.
\newblock In \emph{Proceedings of the 2009 Conference on Empirical Methods in
  Natural Language Processing}. Association for Computational Linguistics.

\bibitem[{Rahman and Ng(2012)}]{rahman2012resolving}
Altaf Rahman and Vincent Ng. 2012.
\newblock Resolving complex cases of definite pronouns: the winograd schema
  challenge.
\newblock In \emph{Proceedings of the 2012 Joint Conference on Empirical
  Methods in Natural Language Processing and Computational Natural Language
  Learning}.

\bibitem[{Rudinger et~al.(2018)Rudinger, Naradowsky, Leonard, and
  Van~Durme}]{rudinger2018gender}
Rachel Rudinger, Jason Naradowsky, Brian Leonard, and Benjamin Van~Durme. 2018.
\newblock Gender bias in coreference resolution.
\newblock In \emph{Proceedings of the 2018 Conference of the North American
  Chapter of the Association for Computational Linguistics: Human Language
  Technologies}.

\bibitem[{Toutanova et~al.(2003)Toutanova, Klein, Manning, and
  Singer}]{toutanova2003feature}
Kristina Toutanova, Dan Klein, Christopher~D Manning, and Yoram Singer. 2003.
\newblock Feature-rich part-of-speech tagging with a cyclic dependency network.
\newblock In \emph{Proceedings of the 2003 Conference of the North American
  Chapter of the Association for Computational Linguistics on Human Language
  Technology}.

\bibitem[{Trichelair et~al.(2018)Trichelair, Emami, Cheung, Trischler, Suleman,
  and Diaz}]{trichelair2018evaluation}
Paul Trichelair, Ali Emami, Jackie Chi~Kit Cheung, Adam Trischler, Kaheer
  Suleman, and Fernando Diaz. 2018.
\newblock On the evaluation of common-sense reasoning in natural language
  understanding.
\newblock \emph{The NeurIPS Workshop on Critiquing and Correcting Trends in
  Machine Learning}.

\bibitem[{Trinh and Le(2018)}]{trinh2018simple}
Trieu~H Trinh and Quoc~V Le. 2018.
\newblock A simple method for commonsense reasoning.
\newblock \emph{arXiv preprint arXiv:1806.02847}.

\bibitem[{Webster et~al.(2018)Webster, Recasens, Axelrod, and
  Baldridge}]{webster2018mind}
Kellie Webster, Marta Recasens, Vera Axelrod, and Jason Baldridge. 2018.
\newblock Mind the gap: A balanced corpus of gendered ambiguous pronouns.
\newblock \emph{Transactions of the Association for Computational Linguistics},
  6:605--617.

\bibitem[{Wiseman et~al.(2016)Wiseman, Rush, and Shieber}]{wiseman2016learning}
Sam Wiseman, Alexander~M Rush, and Stuart~M Shieber. 2016.
\newblock Learning global features for coreference resolution.
\newblock In \emph{Proceedings of NAACL-HLT}.

\bibitem[{Zhao et~al.(2018)Zhao, Wang, Yatskar, Ordonez, and
  Chang}]{zhao2018gender}
Jieyu Zhao, Tianlu Wang, Mark Yatskar, Vicente Ordonez, and Kai-Wei Chang.
  2018.
\newblock Gender bias in coreference resolution: Evaluation and debiasing
  methods.
\newblock In \emph{NAACL-HLT}.

\end{thebibliography}
\bibliographystyle{acl_natbib}

\end{document}